\documentclass[10pt,twocolumn,letterpaper]{article}

\usepackage{iccv}      

\definecolor{iccvblue}{rgb}{0.21,0.49,0.74}
\usepackage[pagebackref,breaklinks,colorlinks,allcolors=iccvblue]{hyperref}

\def\paperID{11637} 
\def\confName{ICCV}
\def\confYear{2025}

\title{ExpFace: Exponential Angular Margin Loss for Deep Face Recognition}

\author{Jinhui Zheng\\
Institution1\\
Institution1 address\\
{\tt\small zhengjinhui95@stu.jnu.edu.cn}
\and
Second Author\\
Institution2\\
First line of institution2 address\\
{\tt\small secondauthor@i2.org}
}

\begin{document}
\maketitle
\begin{abstract}
Face recognition is an open-set problem requiring high discriminative power to ensure that intra-class distances remain smaller than inter-class distances. Margin-based softmax losses, such as SphereFace, CosFace, and ArcFace, have been widely adopted to enhance intra-class compactness and inter-class separability, yet they overlook the impact of noisy samples. By examining the distribution of samples in the angular space, we observe that clean samples predominantly cluster in the center region, whereas noisy samples tend to shift toward the peripheral region. Motivated by this observation, we propose the Exponential Angular Margin Loss (ExpFace), which introduces an angular exponential term as the margin. This design applies a larger penalty in the center region and a smaller penalty in the peripheral region within the angular space, thereby emphasizing clean samples while suppressing noisy samples. We present a unified analysis of ExpFace and classical margin-based softmax losses in terms of margin embedding forms, similarity curves, and gradient curves, showing that ExpFace not only avoids the training instability of SphereFace and the non-monotonicity of ArcFace, but also exhibits a similarity curve that applies penalties in the same manner as the decision boundary in the angular space. Extensive experiments demonstrate that ExpFace achieves state-of-the-art performance. To facilitate future research, we have released the source code at: \url{https://github.com/dfr-code/ExpFace}.
\end{abstract}    
\section{Introduction}
\label{sec:intro}





\begin{figure}[t]
    \centering
    \includegraphics[width=1\linewidth]{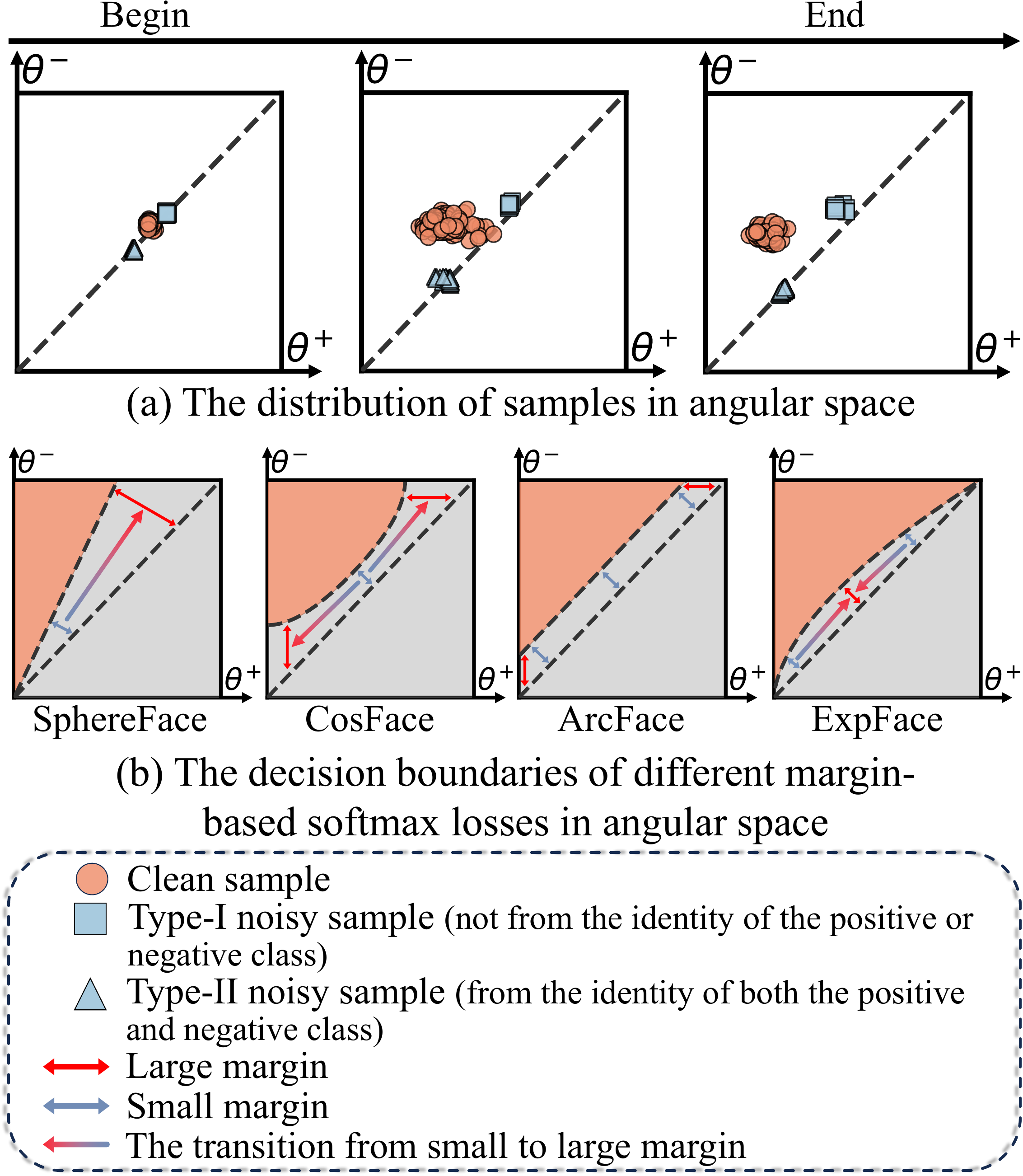}
    \caption{(a) Distribution of different samples in angular space during training: Clean samples are mainly located in the central region. Type-I noisy samples tend to move toward the upper-right boundary region. Type-II noisy samples tend to move toward the lower-left boundary region. (b) Margin distribution of different margin-based softmax losses in angular space: SphereFace applies a margin that increases along $y=x$, from the lower-left to the upper-right.CosFace applies a margin that increases along $y=x$ from the center to the boundary. ArcFace applies a mostly uniform margin, except for larger margins near the endpoints at the upper-right and lower-left corners. The proposed ExpFace applies a margin that increases along $y=x$ from the boundary toward the center.}
    \label{fig:angle space}
\end{figure}
Face recognition constitutes an open-set problem, necessitating that the model possesses high discriminative power to ensure that the distance between samples of the same class remains smaller than that between samples of different classes. However, the traditional softmax, designed for closed-set problems, cannot meet this need. To address this challenge, a mainstream approach involves the use of margin-based softmax losses for model training. This is accomplished by incorporating a margin penalty that constrains the angle between a sample and its corresponding positive class center, thereby enhancing intra-class compactness and inter-class separability. Owing to its simplicity and effectiveness, this method is widely adopted and becomes one of the essential techniques in the design of losses for modern face recognition systems.

Moreover, the advancement of face recognition technology also relies on the expansion of face datasets, where a sufficient volume of facial images ensures the generalization capability of recognition models. Due to labor cost constraints, these large-scale datasets typically undergo automated cleaning procedures. However, this approach cannot fully eliminate all noisy samples. Consequently, mitigating the impact of noisy samples in datasets on model training becomes a key research focus in the field of face recognition.

During training, we observe that clean samples and noisy samples show different distribution patterns in angular space, as illustrated in Fig. \ref{fig:angle space}a. Clean samples are located in the central region. In contrast, noisy samples that are not from the identity of either the positive class or the negative class, referred to as \textbf{Type-I noisy samples}, tend to move toward the upper-right boundary region. Noisy samples that are from the identity of both the positive and negative class, referred to as \textbf{Type-II noisy samples}, tend to move toward the lower-left boundary region. A detailed analysis is provided in Section \ref{ssc:Sample Distribution}. Intuitively, to effectively reduce the influence of noisy samples, margin-based softmax losses should apply stronger penalties to the central region and weaker penalties to the boundary regions. However, existing margin-based softmax losses \cite{liu2016large, liu2022sphereface, liu2017sphereface, wang2018cosface, wang2018additive, deng2022arcface, deng2019arcface} overlook this consideration. As shown in Fig. \ref{fig:angle space}b, different margin-based softmax losses exhibit different decision boundary patterns in angular space. Specifically, SphereFace \cite{liu2016large, liu2022sphereface, liu2017sphereface} imposes a margin that increases along $y=x$, from the lower-left to the upper-right, indicating a stronger focus on samples with large angles to both the positive and negative class centers. CosFace \cite{wang2018cosface, wang2018additive} applies a margin that increases from the center toward both ends along $y=x$, suggesting a focus on boundary samples. ArcFace \cite{deng2022arcface, deng2019arcface} applies a mostly uniform margin, except for larger margins near the endpoints at the upper-right and lower-left corners.


To address this issue, we propose Exponential Angular Margin Loss (ExpFace), a simple yet effective margin-based loss that employs an exponential angular term to penalize samples. Details are provided in Section \ref{ssc:expface}. In angular space, ExpFace applies stronger penalties to samples near the center and progressively weaker penalties to those toward the periphery. This ensures that clean samples receive greater emphasis, while noisy samples are suppressed, which aligns precisely with our intuition.

In addition to analyzing ExpFace and three classic margin-based softmax losses from the perspective of angle space, we also conduct a unified analysis from three aspects: the similarity computation formula, the similarity curve, and the gradient curve. We find that ExpFace has superior properties. 1) The implementation of ExpFace is simple and does not face the training instability  SphereFace. 2) When the embedded margin increases, ExpFace does not encounter the problem of expanding non-monotonic intervals like ArcFace which implies good generalizability. 3) The similarity curve of ExpFace shows large penalties in the center and small penalties at the periphery, which is consistent with the property of the decision boundary in angle space. See Section \ref{ssc:Analysis} for details.

In summary, this paper has the following contributions:
\begin{itemize}
\item We propose a novel margin-based softmax loss called ExpFace, following the design principles of SphereFace, CosFace, and ArcFace. In angle space, ExpFace embeds a large margin in the center region where clean samples are located and a small margin in the peripheral region where few noisy samples are located. This simple yet effective approach mitigates the impact of noisy samples on training and improves training efficiency.
\item We perform a unified analysis of ExpFace and three classic margin-based softmax losses from three aspects: the embedding form of margin, the similarity curve, and the gradient curve. This analysis shows that ExpFace not only remedies the shortcomings of SphereFace and ArcFace, but also the same property on the similarity curve as the decision boundary in angle space, that is imposing large penalties in the center and small penalties at the periphery.
\item We supervise the training of models with ExpFace on the CASIA, MS1MV3, and WebFace4M datasets, and then validate models on popular test sets. The results demonstrate that ExpFace achieves state-of-the-art performance. Refer to Section \ref{sc:experiments} for more information.
\end{itemize}

\section{Related work}
\label{sc:rw}
To enhance the discriminative power of DCNNs for face recognition, two primary methods of loss design have emerged: metric-based methods and softmax-based methods. Specifically, metric-based methods optimize feature distributions by learning similarity metrics between features, explicitly minimizing intra-class distances while maximizing inter-class separability, such as contrastive loss \cite{chopra2005learning} and triplet loss \cite{schroff2015facenet}, but the methods require complex sample mining strategies and incur substantial computational costs. 

In contrast, Softmax-based methods \cite{centerloss, liu2016large, liu2022sphereface, liu2017sphereface, wang2018cosface, wang2018additive, deng2022arcface, deng2019arcface} are free from these limitations and enhance the standard softmax by incorporating additional constraints or reformulations, ensuring the discriminability of features while correctly classifying. Among these, the most popular method is margin-based softmax loss, which first maps the features of samples and class centers to the angular space through $L_{2}$ normalization, and then embeds a margin to penalize the angles between sample features and their corresponding class centers, thereby enhancing discriminative power of the model. Specifically, SphereFace \cite{liu2022sphereface, liu2016large, liu2017sphereface}, CosFace\cite{wang2018additive,wang2018cosface}, and ArcFace \cite{deng2022arcface, deng2019arcface} employ an angular multiplicative margin, a cosine additive margin, and an angular additive margin, respectively.


Recently, thanks to their simplicity and effectiveness, margin-based methods are widely applied in research on losses for face recognition to enhance model performance. For example, \cite{duan2019uniformface} and \cite{zhao2019regularface}, building on the margin-based softmax loss, incorporate regularization terms that explicitly consider class center distributions, effectively improving inter-class separability. \cite{deng2020sub} and \cite{an2022killing} improve margin-based softmax losses by redesigning the number and configuration of class centers. \cite{zhang2019adacos} proposes an adaptive strategy that dynamically adjusts the scaling parameters of CosFace based on the angular distribution of each training batch, thereby optimizing its performance. \cite{boutros2022elasticface} identifies that applying fixed margin penalties to samples across different identities in margin-based softmax loss is suboptimal for real data with inconsistent intra-class and inter-class variations.

As the scale of face datasets grows, noisy samples are hard to clean and disturb the training, but the three classic margin-based softmax losses are designed without regard to such contamination. Inspired by the observation that image quality correlates with the magnitude of the feature, MagFace \cite{magface} and AdaFace \cite{adaface} adjust the margin along the magnitude dimension, assigning distinct margins to clean and noisy samples. CurricularFace \cite{huang2020curricularface} incorporates training progress as an additional dimension to adjust the similarity between different samples and its negative class centers. Circle Loss \cite{CircleLoss} shifts the boundary of the margin-based softmax loss in similarity space so that each sample receives a margin suited to its location, refining training. In this work we reveal that clean and noisy samples occupy different regions of the angular space: clean samples cluster near the center region, while noisy ones drift to the peripheral region. Building on this observation, we propose a margin-based softmax loss, called ExpFace, that imposes large margins on central samples and small margins on peripheral ones, thereby attenuating the impact of noise.

\section{Proposed Approach}
\label{sc:architecture}
\subsection{Margin-based Softmax Loss Revisited}
\label{ssc:margin loss}
The traditional softmax loss, widely adopted for classification tasks, is formulated as follows:
\begin{equation}
\label{eq:Softmax}
L=-\frac{1}{N}\sum^{N}_{i=1}  \ln P_i=-\frac{1}{N}\sum^{N}_{i=1}  \ln\frac{e^{W_{y_{i}}^Tx_{i}+b_{y_{i}}}}{\sum^{C}_{j=1}e^{W_{j}^Tx_{i}+b_{j}}}
\end{equation}
where $N$ is the batch size, $C$ is the number of classes, and $x_{i} \in \mathbb{R}^{d}$ is the feature of the $i$-th sample belong to the $y_{i}$-th class. The feature dimension $d$ usually is $512$, as in \cite{liu2017sphereface, wang2018cosface, deng2019arcface}. $P_i$ denotes the predicted probability that $x_i$ is correctly classified into $y_{i}$--th class. $W \in \mathbb{R}^{d \times C} $ and $b \in \mathbb{R}^{C}$ are the weight matrix and bias of the last fully connected layer, respectively.

Building on Eq. \ref{eq:Softmax}, the margin-based softmax loss first regard $W_{j}$ as the feature representing the class center of the $j$-th identity according to \cite{liu2017sphereface}. Then, based on $W^{T}_{j}x_{i}=\left \| W^{T}_{j} \right \| \left \|x_{i} \right \|\cos{\theta} $, both $\left \| W^{T}_{j} \right \|$ and $ \left \|x_{i} \right \|$ are fixed to 1 through $L_{2}$ normalization. To ensure training efficiency, $ \left \|x_{i} \right \|$ is rescaled by the factor $s$. For simplicity, $b_{j}$ is set to 0. The reformulated form is shown as follows:
\begin{equation}
	\label{eq:r-Softmax}
L= -\frac{1}{N}\sum^{N}_{i=1}  \ln\frac{e^{sT(\theta_{y_{i}i})}}{e^{sT(\theta_{y_{i}i})}+\sum^{C}_{j=1;j\ne y_{i} }e^{s\cos{\theta_{ji}}}}
\end{equation}
\begin{equation}
T(\theta_{y_{i}i})=\cos{\theta_{y_{i}i}}
\label{eq:T}
\end{equation}
where $\theta_{ji}$ represents the angle between $W_{j}$ and $x_{i}$, and $T(\theta_{y_{i}i})$ denotes the similarity computation formula for computing the similarity between $x_{i}$ and its positive class center $W_{y_{i}i}$.
The reformulated form maps the features and class centers into the angular space, thereby explicitly optimizing the angle between features and class centers during the training process. 

Furthermore, the margin-based softmax loss penalizes $\theta_{ji}$ by embedding a margin into $T(\theta_{y_{i}i})$. Specifically, the three classic margin-based softmax losses, SphereFace \cite{liu2022sphereface, liu2016large, liu2017sphereface}, CosFace \cite{wang2018additive,wang2018cosface} and ArcFace \cite{deng2022arcface, deng2019arcface}, embed a margin into $T(\theta_{y_i i})$ by applying a multiplicative angular term, an additive cosine term and an additive angular term respectively, as shown below.
\begin{equation}
T_{SphereFace}(\theta_{y_{i}i})=(-1)^{k}\cos{(m_{s}\theta_{y_{i}i})}-2k
\label{eq:Tsphereface}
\end{equation}
\begin{equation}
T_{CosFace}(\theta_{y_{i}i})=\cos{\theta_{y_{i}i}}-m_{c}
\label{eq:Tcosface}
\end{equation}
\begin{equation}
T_{ArcFace}(\theta_{y_{i}i})=\cos{(\theta_{y_{i}i}+m_{a})}
\label{eq:Tarcface}
\end{equation}
where the hyper-parameter $m_{s}$, $m_{c}$, and $m_{a}$ denote the margin of SphereFace, CosFace, and ArcFace, respectively. Eq. \ref{eq:Tsphereface} is subject to the constraint $\theta \in [\frac{k\pi}{m},\frac{(k+1)\pi}{m}], k\in\mathbb{N}$, to ensure the monotonicity of $T_{SphereFace}(\theta_{y_{i}i})$. Typically, $m_{s}=1.7$, $m_{c}=0.4$, and $m_{a}=0.5$ reduce $T(\theta_{y_{i}i})$ and raise $L$. so a smaller $\theta_{y_{i}i}$ is required to match the loss without the margin, resulting in more discriminative features under their supervision.
\subsection{Sample Distribution in Angular Space}
\label{ssc:Sample Distribution}

\begin{figure}[t]
    \centering
    \includegraphics[width=1\linewidth]{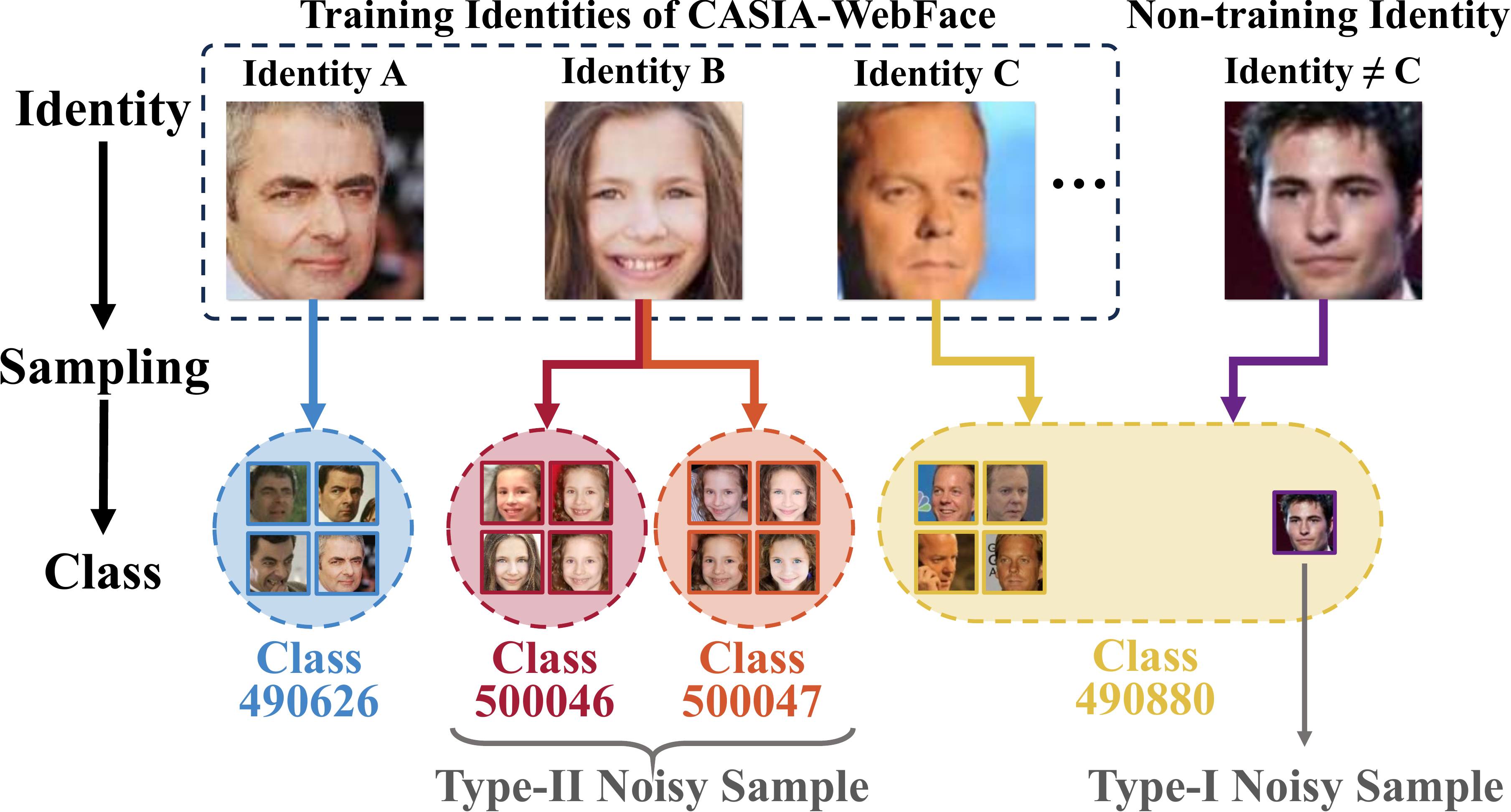}
    \caption{Example of the training set construction process using the CASIA-WebFace dataset \cite{yi2014learning}. Multiple identities are selected as training identities, and face images are sampled from them, with samples from the same identity stored in the same class (folder). Normally, each training identity is assigned to only one class (e.g., Identity A to Class 490626). However, a non-training identity may be sampled into a class (e.g., Identity $\ne$ C to Class 490880), or a training identity may be sampled into two different classes (e.g., Identity B to Class 500046 and Class 500047). These cases result in Type-I and Type-II noisy samples, respectively.}
    \label{fig:sampling}
\end{figure}
Fig. \ref{fig:sampling} illustrates the construction process of the training set. In this process, multiple identities are selected as training identities, and face images are sampled from each identity to form corresponding classes. However, during the sampling stage, two types of noisy cases may occur: 1) Face images from non-training identities may be mistakenly sampled and assigned to a class that belongs to a training identity. We refer to such images as Type-I noisy samples. During training, because the identity of the sample does not match the assigned class (positive class), the similarity between the sample and the positive class center decreases, resulting in a larger angle. At the same time, since it also does not belong to any negative class identity, its similarities with the negative class centers also decrease, and the angles increase. As a result, in angular space, Type-I noisy samples tend to move toward the upper-right boundary region. 2) Face images from a training identity may be repeatedly sampled and incorrectly assigned to multiple different classes. These images are referred to as Type-II noisy samples. Since all these classes correspond to the same identity, as training progresses, the similarities between these samples and the class centers of both the positive and negative classes within these constructed classes increase, leading to smaller angles. As a result, in angular space, Type-II noisy samples tend to move toward the lower-left boundary region.

Overall, we observe that clean samples predominantly reside in the central region throughout training, while noise samples drift toward the periphery region. Thus, an effective strategy to mitigate noise samples is to impose larger penalties in the central region and smaller penalties in the peripheral region, thereby focusing on clean samples while ignoring noise samples. However, the three classic margin-based softmax losses overlook this, as shown in Fig. \ref{fig:angle space}(b). In angular space, the margin of SphereFace increases along $y=x$ from the lower-left to the upper-right periphery region, the margin of CosFace rises along $y=x$ from the central region to the periphery regions. the margin of ArcFace is mostly uniform, with slightly larger margins near the lower-left and upper-right endpoints. Inspired by the above observation, we propose a new margin-based softmax loss, called ExpFace, which is simple but effective in mitigating the impact of noise samples.

\begin{figure*}[htbp]
  \centering
  \begin{subfigure}{0.24\linewidth}
    \includegraphics[width=1\textwidth]{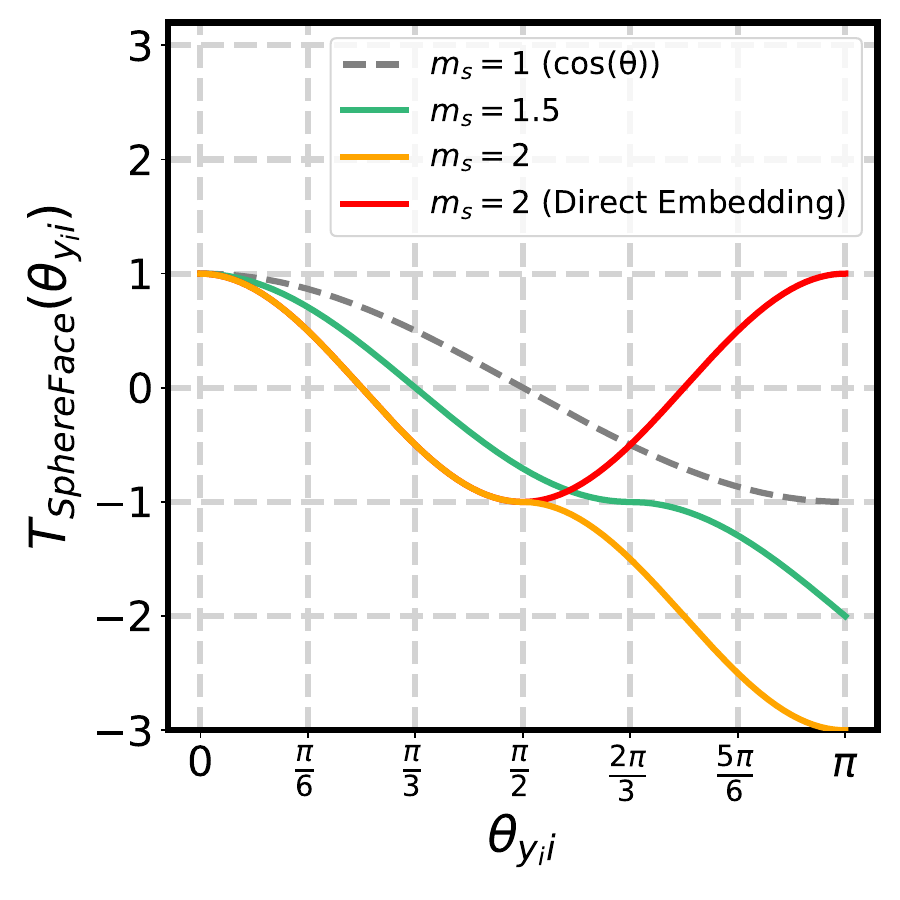}
    \caption{SphereFace}
    \label{sfig:sphereface_t}
  \end{subfigure}
  \begin{subfigure}{0.24\linewidth}
    \includegraphics[width=1\textwidth]{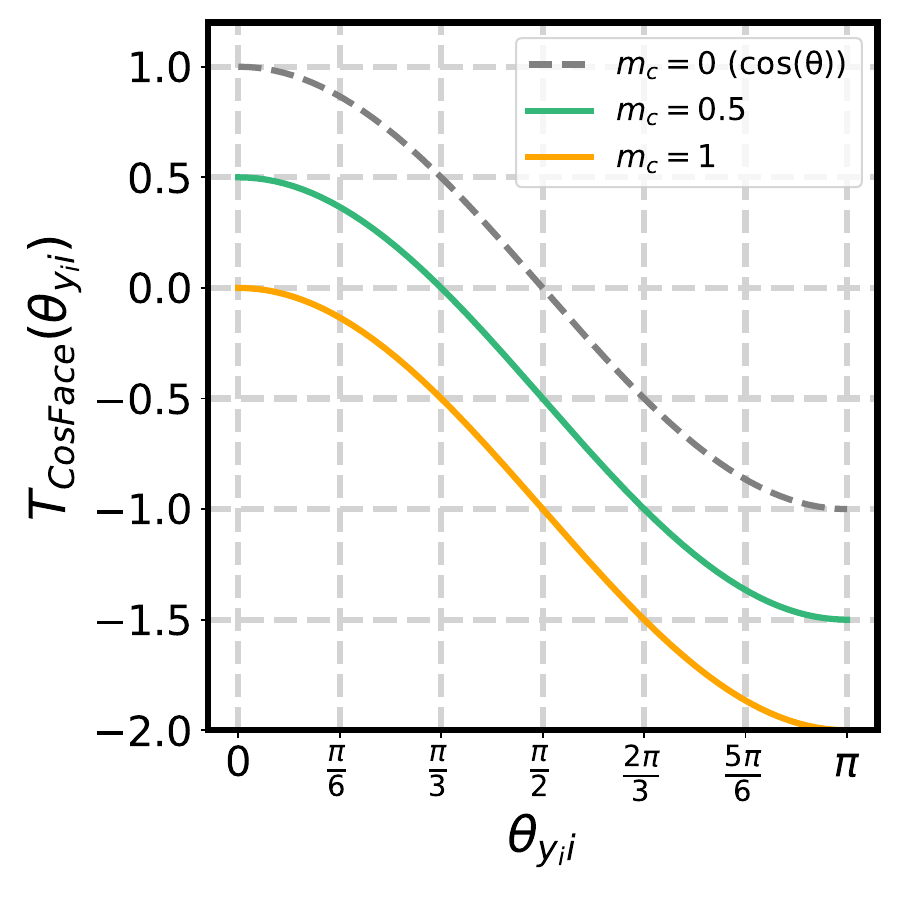}
    \caption{CosFace}
    \label{sfig:cosface_t}
  \end{subfigure}
  \begin{subfigure}{0.24\linewidth}
    \includegraphics[width=1\textwidth]{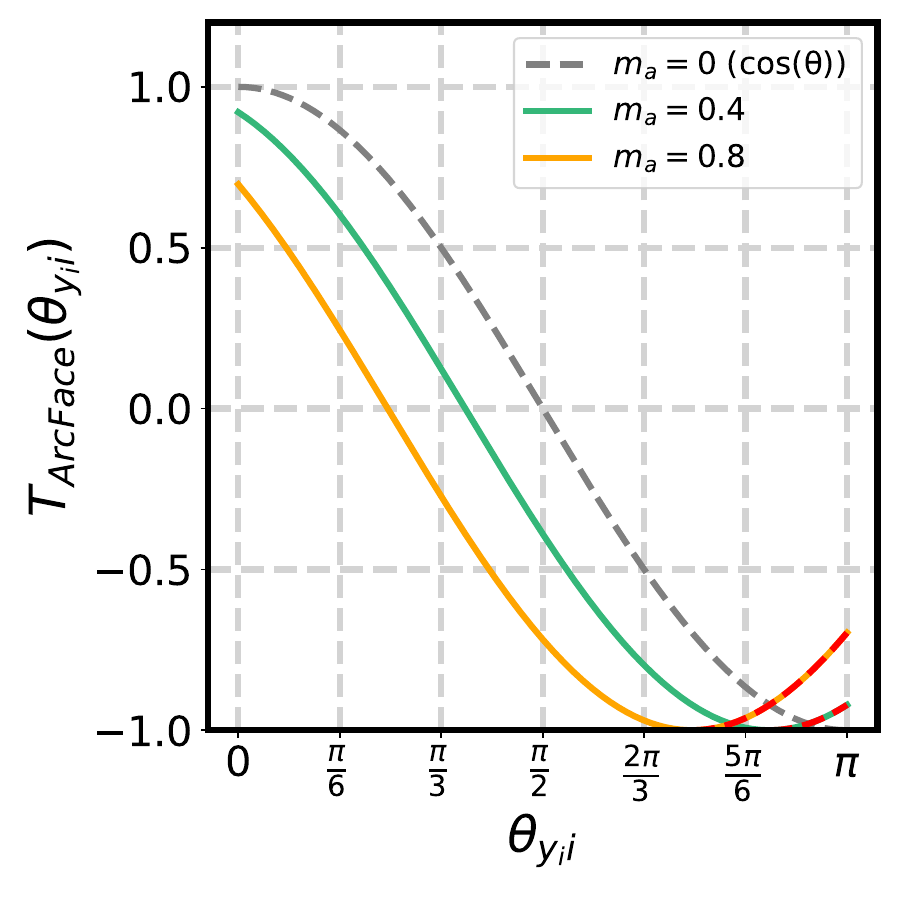}
    \caption{ArcFace}
    \label{sfig:arcface_t}
  \end{subfigure}
  \begin{subfigure}{0.24\linewidth}
    \includegraphics[width=1\textwidth]{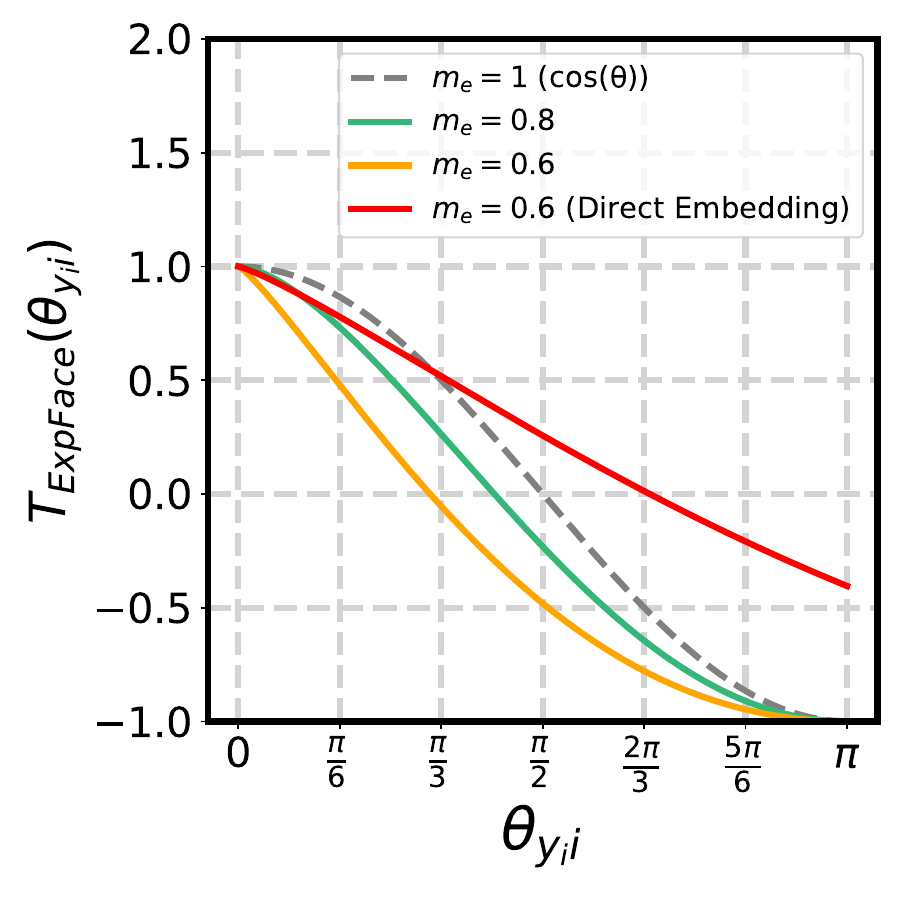}
    \caption{ExpFace}
    \label{sfig:expface_t}
  \end{subfigure}
  \caption{Similarity curves between sample features and their corresponding positive class centers for different margin-based softmax losses under various margin settings. The gray dashed line represents the similarity curve without embedding a margin. In (a) and (b), the red solid lines show the similarity curves of SphereFace and ExpFace when the margin is directly embedded. In (c), the red dashed line highlights the monotonic increasing part of the ArcFace similarity curve.}
  \label{fig:T}
\end{figure*}

\subsection{ExpFace}
\label{ssc:expface}

Following the design principles of SphereFace, CosFace, and ArcFace, this paper proposes to embed an exponential angular term as the fourth margin to $T(\theta_{y_{i}i})$, thereby penalizing the angle between features and positive class centers to enhance the discriminative power of the trained model. The implemented form is shown as follows: 
\begin{equation}
	\label{eq:T_exp_1}
T_{ExpFace}(\theta_{y_{i}i})=\cos{(\theta_{{y_{i}i}}^{m_e})}
\end{equation}
whose curve is shown as the red line in Fig. \ref{sfig:expface_t}. It is observed that when the margin is directly embedded, the left part of the curve lies below the dashed gray line, while the right part lies above it. This indicates that the margin cannot effectively penalize samples with large $\theta_{y_{i}i}$. To resolve this issue, we first shrink the $\theta_{y_{i}i}$ values by dividing $\pi$, subsequently embed the margin to impose the penalty, and finally rescale the values by multiplying by $\pi$. The novel form is formulated as follows: 
\begin{equation}
	\label{eq:T_exp}
T_{ExpFace}(\theta_{y_{i}i})=\cos{(\pi(\frac{\theta_{{y_{i}i}}}{\pi} )^{m_e})}
\end{equation}

Compared with the three classical margin-based softmax losses, ExpFace imposes larger penalties on samples in the central region and smaller penalties on samples in the peripheral region in the angular space, thereby focusing model training on clean samples while ignoring noise samples, which improves training efficiency.


\subsection{Unified Analysis of Margin-based Softmax Losses}
\label{ssc:Analysis}
In this section, we conduct a unified analysis of SphereFace, CosFace, ArcFace, and the proposed ExpFace from three perspectives: the similarity computation formula, the similarity curve, and the gradient curve. This helps provide a deeper understanding of the properties of ExpFace.

\noindent\textbf{Analysis of the Similarity Computation Formula.} By examining the similarity computation formula of three representative margin-based loss functions, as shown in Eq. \ref{eq:Tsphereface}, Eq. \ref{eq:Tcosface}, and Eq. \ref{eq:Tarcface}, we observe that SphereFace employs a rather complex form to ensure the monotonicity is preserved before and after embedding margin. However, this design leads to high computational complexity and results in training instability. In contrast, the proposed ExpFace adopts the similarity computation formula as shown in Eq. \ref{eq:T_exp_1}, where an angular exponential term $m_e$ is applied. It employs a \textbf{shrink-then-expand} design that ensures effective penalization across the entire range of $\theta_{y_ii}\in [0,\pi]$. This formula is simple and does not introduce significant additional computational cost, thereby maintaining training stability.

\noindent\textbf{Analysis of the Similarity Curve.} According to the four margin-based softmax losses, we plot the similarity curves between the samples and their corresponding positive class centers, as shown in Fig. \ref{fig:T}. From an overall perspective, we can observe that embedding a margin in the $T(\theta_{y_ii})$ results in the curve lying below the gray dashed line, which represents Eq. \ref{eq:T}. In addition, a larger margin causes the entire curve to lie further below. Next, we analyze the characteristics of each similarity curve individually. First, from Fig. \ref{sfig:sphereface_t}, we can see that directly using a multiplicative angular term as the margin causes the similarity curve to fail to apply effective penalties at large values of $\theta_{y_ii}$, as shown by the red solid line. Therefore, SphereFace designs its $T(\theta_{y_ii})$ specifically to ensure that the curve stays below the gray dashed line. Second, from Fig. \ref{sfig:cosface_t}, we observe that the similarity curve of CosFace shifts downward as the margin $m_c$ increases. This indicates that CosFace applies a consistent penalty on the similarity between samples and their positive class centers across all $\theta_{y_ii}$ values. Third, from Fig. \ref{sfig:arcface_t}, we see that as the margin $m_a$ increases, the similarity curve of ArcFace shifts to the left. At the same time, the monotonic increasing region on the right side of the curve becomes wider, as shown by the red dashed line. This reduces the generalizability of ArcFace. Fourth, from Fig. \ref{sfig:expface_t}, we can see that, as shown by the red solid line, directly embedding an angular exponential term, similar to SphereFace, fails to impose effective penalties at large values of $\theta_{y_ii}$. Therefore, its $T(\theta_{y_ii})$ must be carefully designed to correct this limitation. The resulting similarity curve not only ensures that it remains below the gray dashed line, but also avoids the non-monotonicity problem seen in ArcFace. In addition, the curve applies stronger penalties in the center region and weaker penalties in the peripheral region. This behavior is consistent with its decision boundary in angular space.

\begin{figure*}
  \centering
  \begin{subfigure}{0.24\linewidth}
    \includegraphics[width=1\textwidth]{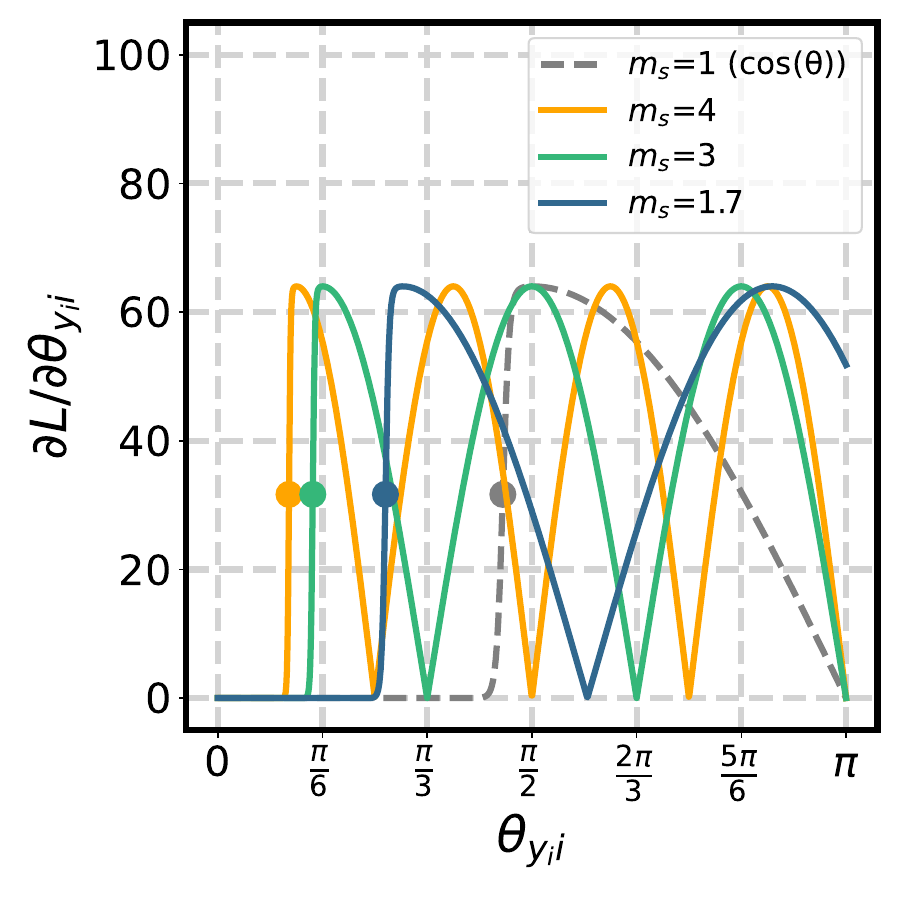}
    \caption{SphereFace}
    \label{sfig:sphereface_GL}
  \end{subfigure}
  \begin{subfigure}{0.24\linewidth}
    \includegraphics[width=1\textwidth]{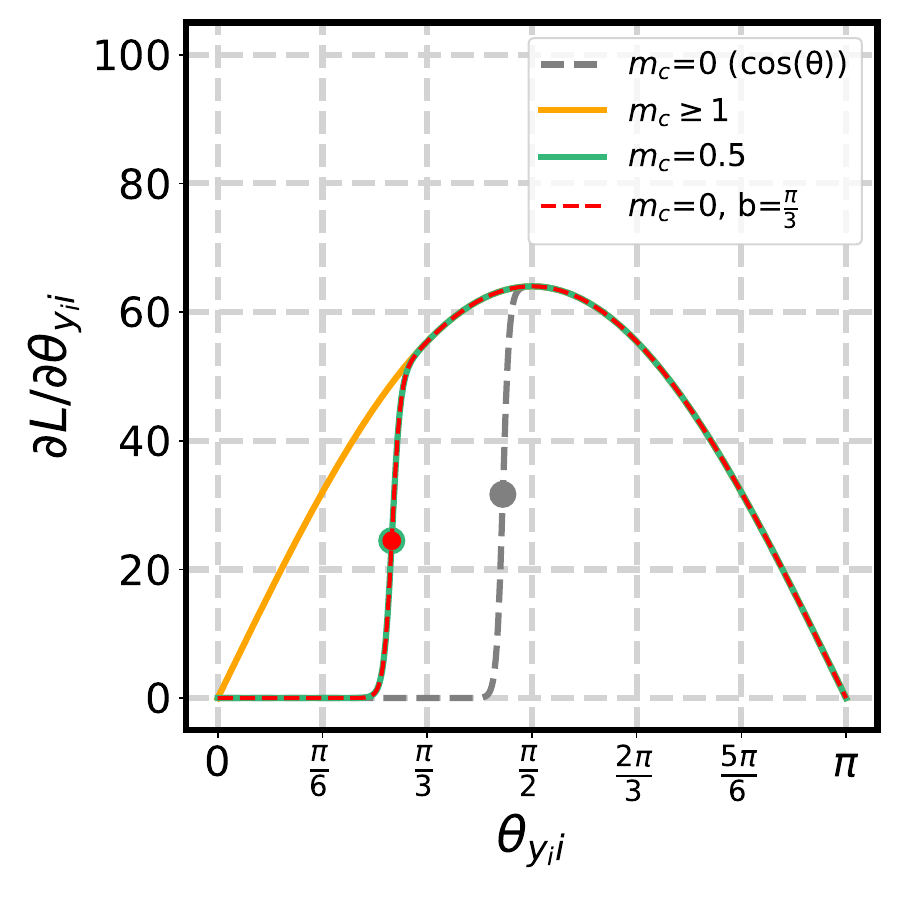}
    \caption{CosFace}
    \label{sfig:cosface_GL}
  \end{subfigure}
  \begin{subfigure}{0.24\linewidth}
    \includegraphics[width=1\textwidth]{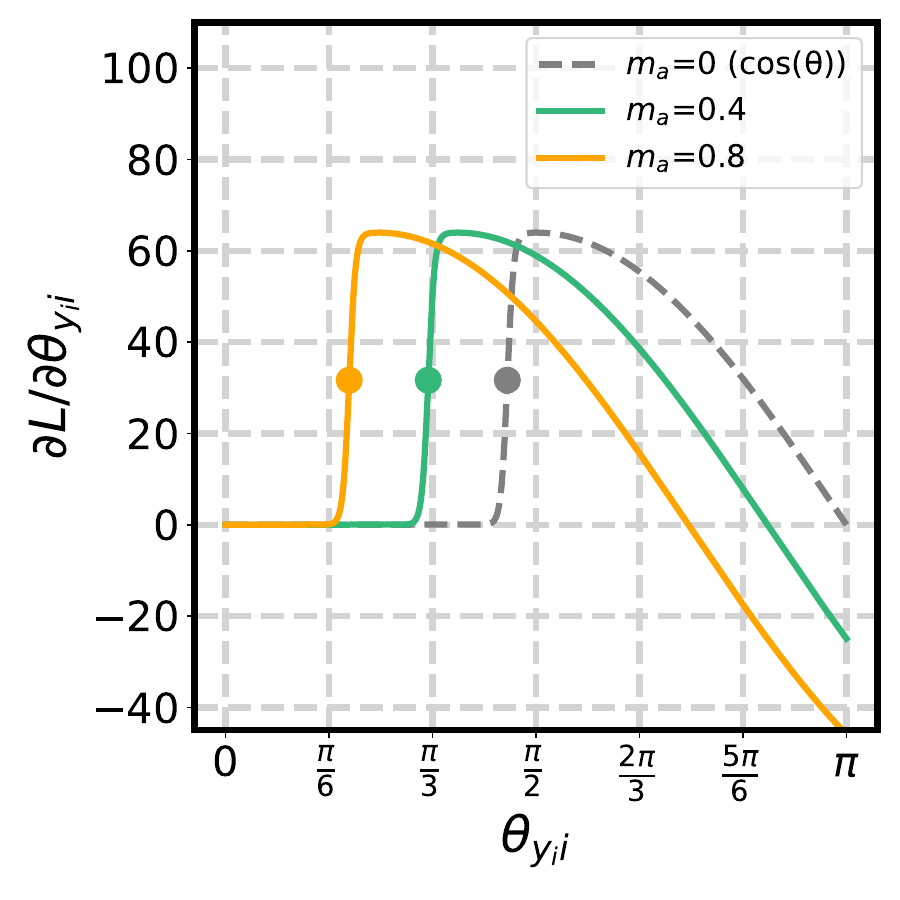}
    \caption{ArcFace}
    \label{sfig:arcface_GL}
  \end{subfigure}
  \begin{subfigure}{0.24\linewidth}
    \includegraphics[width=1\textwidth]{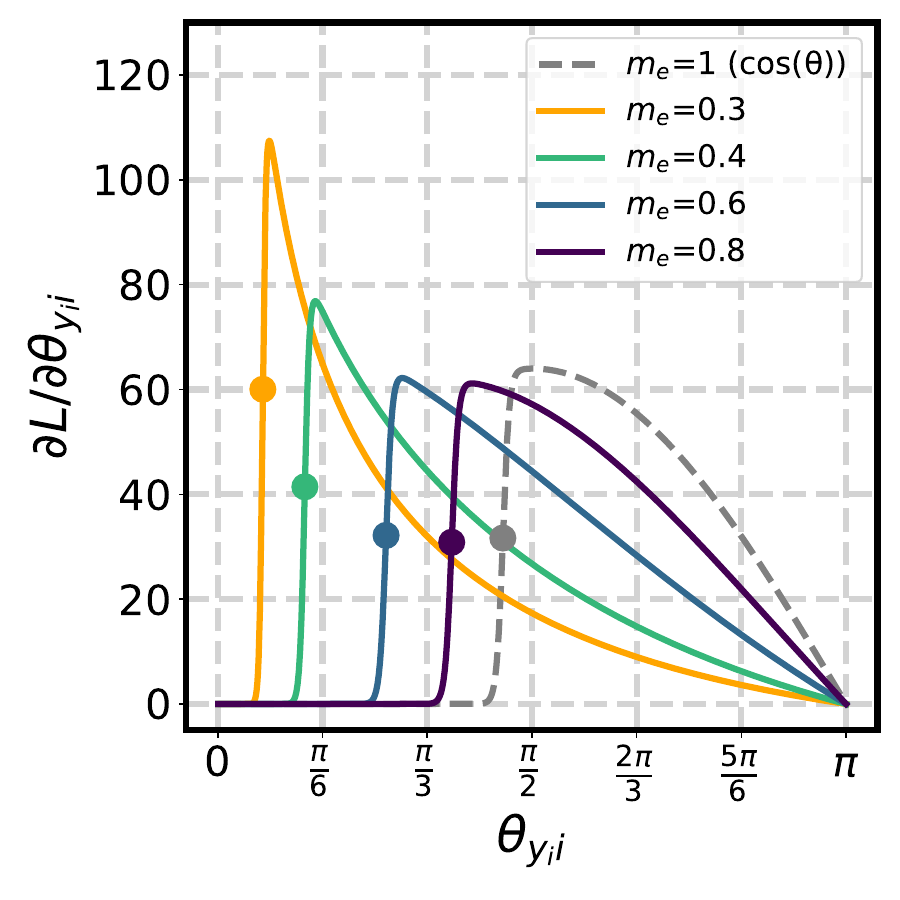}
    \caption{ExpFace}
    \label{sfig:expface_GL}
  \end{subfigure}
  \caption{Gradient curves of different margin-based softmax losses under various margin settings. The gray dashed line represents the gradient curve without embedding a margin. Solid dots mark points in the region where the gradient changes rapidly, with their positions computed as described in Section \ref{ssc:Analysis} on \textbf{Analysis of the Gradient Curve}.}
  \label{fig:G}
\end{figure*}

\noindent\textbf{Analysis of the Gradient Curve.} During the training process, the backpropagation stage updates the model parameters based on the gradients to reduce $L$. Consequently, the gradient of $L$ w.r.t $\theta_{y_{i}i}$ plays a pivotal role throughout the training. Building on this, we compares ExpFace with three classical margin-based softmax losses from the perspective of this gradient to investigate its properties. For ease of analysis, we set $N$ to 1 and take the average of the angles between sample feature $x_i$ and all negative class centers $\{W_{j} | j\ne y_i\}$ as $b$. We then further derive Eq. \ref{eq:r-Softmax}, as demonstrated below:
\begin{equation}
\begin{aligned}
\label{eq:r-Softmax-b}
L &=-\ln\frac{e^{sT(\theta_{y_{i}i})}}{e^{sT(\theta_{y_{i}i})}+(C-1)e^{s\cos{(b)}}}\\
&=-\ln\frac{e^{sT(\theta_{y_{i}i})}}{e^{sT(\theta_{y_{i}i})}+e^{s\cos{(b)}+\ln{(C-1)}}}\\
&=\ln{(1+e^{-sT(\theta_{y_{i}i})+s\cos{(b)}+\ln(C-1)})}
\end{aligned}
\end{equation}
We illustrate the gradient curves of $L$ w.r.t. $\theta_{y_{i}i}$ for different margin-based softmax losses across various values of $m$, based on the derivation outcome. Specifically, following the observation in \cite{zhang2019adacos} that $b$ stabilizes at approximately $\frac{\pi}{2}$ during training, we set $b$ to $\frac{\pi}{2}$ unless specifically indicated otherwise. Concurrently, we assign the value of $C$ to $10573$, which denotes the number of identities in the CASIA-WebFace dataset \cite{yi2014learning}. The results are depicted in Fig. \ref{fig:G}. 

We can observe that the gradient curve without embedding a margin, represented by the gray dashed line in Fig. \ref{fig:G}, contains a region where the gradient changes rapidly. When $\theta_{y_{i}i}$ is located to the left of this region, the loss converges due to the gradient approaches $0$. Furthermore, we find that $\theta_{trans}$ lies within this region when it satisfies the equation $P_i=\frac{e^{sT(\theta_{y_{i}i})}}{e^{sT(\theta_{y_{i}i})}+e^{s\cos{(b)}+\ln{(C-1)}}}=\frac{1}{2}$. Therefore, we refer to this $\theta_{trans}$ as the \textbf{transition angle}. We further derive this as follows:
\begin{equation}
\label{eq:traistion-1}
        \frac{e^{sT(\theta_{y_{i}i})}}{e^{sT(\theta_{y_{i}i})}+e^{s\cos{(b)}+\ln{(C-1)}}}=\frac{1}{2}
\end{equation}
\begin{equation}
\label{eq:traistion-2}
        e^{sT(\theta_{y_{i}i})}=e^{s\cos{(b)}+\ln{(C-1)}}
\end{equation}
\begin{equation}
\label{eq:traistion-3}
        sT(\theta_{trans})=s\cos{b}+\ln{(C-1)}
\end{equation}
\begin{equation}
\label{eq:traistion-4}
        T(\theta_{trans})=\cos{b}+\frac{\ln{(C-1)}}{s}
\end{equation}
\begin{equation}
\label{eq:traistion-5}
    \theta_{trans}=T^{-1}(\cos{b}+\frac{\ln{(C-1)}}{s})
\end{equation}
By substituting Eq. \ref{eq:T} into the derivation outcome, we can determine that $\theta_{trans}=\arccos{(\cos{b}+\frac{\ln{(C-1)}}{s})}$. Since we set $b=\frac{\pi}{2}$, we can easily deduce that the transition angle is slightly less than $\frac{\pi}{2}$. This implies that at $\theta_{trans}$, the samples are just enough to ensure correct classification to the positive class. Additionally, the gray dashed line attains its maximum at $\theta_{max}=\frac{\pi}{2}$, indicating that samples near this angle have the most significant impact on model parameters. Consequently, the learning focus can be considered to be on these samples. 

Based on Eq. \ref{eq:traistion-5}, we compute $\theta_{trans}$ for all gradient curves in Fig. \ref{fig:G} and mark the corresponding positions with solid dots. From these points, we observe that all margin-based softmax losses shift $\theta_{trans}$ to the left by adjusting the margin. This makes training stricter, as a smaller $\theta_{y_ii}$ is required for the loss to converge, thereby improving feature discriminability. 

In addition to the common characteristics above, the following summarizes the unique features of the gradient curves of different margin-based softmax losses. For SphereFace, as illustrated in Fig. \ref{sfig:sphereface_GL}, a new peak emerge in the gradient curve when $m_s\ge1.5$, and as $m_s$ increases, the number of both peaks and troughs in the gradient curve progressively increases. This results in a decrease in the contribution of more samples to the training of the model, as they are situated near the troughs. Furthermore, the presence of multiple peaks results in a larger number of significant gradients that offer update directions, thus reducing the stability of the training process.

In the case of CosFace, Fig. \ref{sfig:cosface_GL} shows that the gradient curve remains constant when $m_c\ge1$, whereas a $\theta_{trans}$ appears when $m_c<1$, causing the right side of the curve to stay unchanged while the left side becomes a horizontal line at $x=0$.
By comparing the gradient curve for $m_c=0.5,b=\frac{\pi}{2}$ (green solid line) with that for $m_c=0,b=\frac{\pi}{3}$ (red dashed line), we can clearly see that $m_c$ and $b$ serve a similar function. A smaller $b$ shifts $\theta_{trans}$ to the left, making the training stricter. Additionally, the maximum of the gradient curve consistently occurs at $\frac{\pi}{2}$ which means that the learning focus cannot be adjusted by modifying $m_c$ and remains on samples near $\theta_{y_{i}i=\frac{\pi}{2}}$, which potentially impacts the application of CosFace in dynamic margin strategies.

Regarding ArcFace, Fig. \ref{sfig:arcface_GL} demonstrates that that the gradient curve shifts to the left as $m_a$ increases. Furthermore, we can observe that a portion of the gradient curve lies below $x=0$, and this portion becomes more pronounced as $m_a$ increases. This causes the $\theta_{y_ii}$ of the samples in this region to increase, which contradicts the training objective. To avoid this issue, ArcFace must refrain from using large $m_a$ values, which in turn reduces its generalizability.

Compared to the three classic margin-based softmax losses, ExpFace addresses their respective limitations. Specifically, as shown in Fig. \ref{sfig:expface_GL}: first, ExpFace does not suffer from the increasing oscillation in the gradient curve observed in SphereFace as $m_e$ increases. Second, the maximum value of the gradient curve changes in a regular manner with $m_e$, allowing the margin to be adjusted to control the current learning focus, which aligns well with dynamic margin strategies. Third, ExpFace does not exhibit the problem of negative gradients that can occur in ArcFace.

Overall, in the theoretical analysis of the three aspects discussed above, ExpFace demonstrates highly favorable properties. It is worth noting that in practical use, $m_e$ should not be set too small or too large, as this may lead to gradient explosion. However, $m_e$ values between 0.3 and 10 are sufficient for most applications and do not cause gradient issues.

\begin{table*}[!t] \normalsize
\centering
\renewcommand\arraystretch{1.25}
	\caption{Verification results (\%) of different margin-based softmax losses on the six benchmark datasets}
	\label{tb:comparative-soa}

	\begin{tabular}{l|l|cccccc}
		\hline
		 \textbf{Method}&Dataset&LFW&CPLFW&CALFW&CFP-FP & AgeDB-30&VGG2-FP\\
   \hline
 SphereFace \cite{liu2017sphereface}&CASIA&\textbf{99}
&85.717
&91.167
&92.814
& 90.3
&89.26
\\
 CosFace \cite{wang2018cosface}&CASIA&98.917
&84.433
&92.05
&89.857
& 91.717
&86.68
\\
 ArcFace \cite{deng2022arcface}&CASIA&98.783
&85.033
&92.35
&90.6
& 91.667
&87.52
\\
 \hline
 ExpFace (our)&CASIA&98.783&
\textbf{86.517}&\textbf{92.767}&\textbf{93.057}& \textbf{92.383}&\textbf{89.4}\\
 \hline
 SphereFace
& MS1MV3& \textbf{99.817}& \textbf{93.2}& 96.2& \textbf{98.577}& 98.133&\textbf{95.76}\\
 CosFace
& MS1MV3& \textbf{99.817}& 92.75& 96.017& 98.529& 98.1&95.4\\
 ArcFace
& MS1MV3& 99.783& 93.083& 96.1& 98.529& \textbf{98.233}&95.4\\
\hline
 ExpFace& MS1MV3& 99.733& 93.117& \textbf{96.233}& 98.414& 98.033&95.46\\
 \hline
 SphereFace
& WebFace4M& \textbf{99.8}
& \textbf{94.25}
& 95.917
& 99.1
& 97.667
&\textbf{96.34}\\
 
CosFace& WebFace4M& 99.767
& 94.183
& 95.967
& 99
& 97.733
&95.82
\\
 ArcFace& WebFace4M& 99.767
& 94
& \textbf{96.017}
& 98.9
& 97.65
&96.2
\\
 \hline
ExpFace& WebFace4M& 99.783
& \textbf{94.25}
& 95.817
& \textbf{99.129}
& \textbf{97.8}
&96.14
\\
\hline
	\end{tabular}
 
\end{table*}

\section{Experiments}
\label{sc:experiments}
This section is divided into four parts: Section \ref{ssc:ec} introduces the experimental conditions.
Section \ref{ssc:csoa} demonstrates that ExpFace achieves state-of-the-art performance through comparison with other margin-based softmax losses.

\subsection{Experimental Settings}
\label{ssc:ec}

For the training datasets, we employ three distinct datasets: CASIA \cite{yi2014learning}, MS1MV3 \cite{deng2019lightweight}, and WebFace4M \cite{zhu2021webface260m}. The CASIA dataset comprises approximately 0.5M images across 10K identities. MS1MV3, a refined version of MS1MV0 obtained through a semi-automated cleaning process, contains 5.1M images representing 91K identities. WebFace4M, a subset of WebFace260M, includes 4M images from 200K identities, collected and cleaned using automated procedures from internet sources.
For the validation sets, we utilize six benchmark datasets: LFW \cite{huang2008labeled}, CPLFW \cite{zheng2018cross}, CALFW \cite{zheng2017cross}, CFP-FP \cite{sengupta2016frontal}, AgeDB-30 \cite{moschoglou2017agedb}, and VGG2-FP \cite{cao2018vggface2}, which encompass variations in age, pose, and background. Additionally, we utilize large-scale datasets such as IJB-B \cite{whitelam2017iarpa}, IJB-C \cite{maze2018iarpa} and MegaFace \cite{kemelmacher2016megaface} for testing. 

\begin{table*}[!t] \small
\centering
\renewcommand\arraystretch{1.25}
	\caption{The 1:1 verification accuracy (TAR@FAR) of different margin-based softmax losses on IJBB and IJBC}
	\label{tb:ijb-soa}
  
	\begin{tabular}{l|l|cccccc|cccccc}
		\hline
		 \textbf{Method}&Dataset&\multicolumn{6}{c|}{IJB-B} & \multicolumn{6}{c}{IJB-C}\\
 & & 1e-6& 1e-5& \textbf{1e-4}& 1e-3& 1e-2&1e-1& 1e-6& 1e-5& \textbf{1e-4}& 1e-3& 1e-2&1e-1\\
   \hline
 SphereFace \cite{liu2017sphereface}&CASIA&\textbf{36.08}&53.58&72.69&85.42& 93.16& 97.71
& 50.17& 64.55& 78.38& 88.43& 94.51&98.16
\\
 CosFace \cite{wang2018cosface}&CASIA&30.91&62.35&75.45&85.45& 92.82& 97.45
& 57.96& 69.72& 80.01& 88.65& 94.44&98.03
\\
 ArcFace \cite{deng2022arcface}&CASIA&22.56&60.44&74.77&85.56& 93.13& 97.63
& 56.16& 68.86& 79.96& 88.31& 94.46&98.03
\\
 \hline
 ExpFace (our)&CASIA&33.41&\textbf{64.92}&\textbf{77.62}&\textbf{87.36}& \textbf{93.72}& \textbf{97.76}
& \textbf{60.04}& \textbf{72.23}& \textbf{82.08}& \textbf{90.01}& \textbf{95.12}&\textbf{98.2}
\\
 \hline
 SphereFace
& MS1MV3& 40.11& 90.93& 95.1& 96.77& \textbf{98.09}& \textbf{98.84}
& 90.67& 94.21& 96.39& 97.73& \textbf{98.62}&\textbf{99.23}
\\
 CosFace
& MS1MV3& 40.19& 90.67& 95.32& \textbf{96.85}& 97.86& 98.72
& 
90.54& 94.95& 96.68& 97.8& 98.49&99.11
\\
 ArcFace
& MS1MV3& 37.77& 91.3& 95.34& 96.8& 97.84& 98.67
& 90.03& \textbf{95.03}& 96.66& 97.81& 98.5&99.09
\\
\hline
 ExpFace& MS1MV3& \textbf{40.57}& \textbf{91.45}& \textbf{95.37}& \textbf{96.85}& 97.83& 98.7
& 
\textbf{91.31}
& 95.02& \textbf{96.69}& \textbf{97.83}& 98.52&99.13
\\
 \hline
 SphereFace
& WebFace4M& \textbf{44.41}& 90.29& 94.76& 96.77& \textbf{98.23}& \textbf{99.2}
& 88.59& 93.99& 96.67& 97.96& \textbf{98.83}&\textbf{99.47}
\\
 
CosFace& WebFace4M& 42.08& 91.37& 95.06& \textbf{96.88}& 97.94& 98.93
& 
89.21& 94.83& 96.8& 98& 98.64&99.3
\\
 ArcFace& WebFace4M& 42.81& \textbf{91.6}& 95.19& \textbf{96.88}& 97.86& 98.85
& \textbf{90.86}& \textbf{95.07}& \textbf{96.95}& 98& 98.64&99.25
\\
 \hline
ExpFace& WebFace4M& 42.87& 91.34& \textbf{95.23}& 96.85& 97.95& 98.95
& 90.84& 94.88& 96.91& \textbf{98.01}& 98.67&99.33
\\
\hline

	\end{tabular}
 
\end{table*}

For pre-processing, we generate the normalised face crops (112 × 112) by utilizing five facial points, following \cite{deng2022arcface, wang2018cosface, liu2017sphereface}. We employ ResNet50 \cite{han2017deep, he2016deep} as the embedding network to extract 512-D face embedding features. Subsequently, we supervise the model training using different margin-based softmax losses as the loss functions.


For the training configuration, we utilize the PyTorch framework \cite{paszke2017automatic} and integrate the apex package to enable mixed-precision training, which reduces memory consumption and improves training speed. We employ the SGD optimizer with an initial learning rate of $0.02$, a momentum of $0.9$, and a weight decay of $5e-4$. The training was conducted on a single GPU for $20$ epochs with a mini-batch size of $256$. During the testing stage, we retain only the embedding network to generate a 512-dimensional feature representation for the normalized input faces.

For hyperparameters, SphereFace, CosFace and ArcFace use $[s=32, m=1.7]$, $[s=64, m=0.4]$, and $[s=64, m=0.5]$. The proposed ExpFace employs $s=64$ following \cite{wang2018cosface, deng2019arcface}, and sets $m=0.7$, which is selected through hype-rparameter tuning experiments.

\subsection{Comparison with State-of-the-Art Margin-based Softmax Losses}
\label{ssc:csoa}
In this section, we will supervise the training of ResNet50 models using SphereFace, CosFace, ArcFace, and the proposed ExpFace on the three training sets under the same training configuration, and then comprehensively compare the performance of these models across various datasets. The validation results are shown in Tab. \ref{tb:comparative-soa}, Tab. \ref{tb:ijb-soa}, and Tab. \ref{tb:megaface}.

Tab. \ref{tb:comparative-soa} reports the validation accuracy of different margin-based softmax losses on six benchmark datasets. On the CASIA training set, ExpFace achieves performance close to the other three margin-based softmax losses. Furthermore, on larger-scale training sets, ExpFace demonstrates superior performance. Specifically, on the MS1MV3 training set, ExpFace achieves the best accuracy on CALFW and second-best accuracy on CPLFW and VGG2-FP. On the WebFace4M training set, ExpFace attains the best accuracy on CPLFW, CFP-FP, and AGEDB-30, and second-best accuracy on LFW, again only behind SphereFace.

Tab. \ref{tb:ijb-soa} presents the True Accept Rate (TAR) of different margin-based softmax losses in the 1:1 verification tasks of IJB-B and IJB-C at various False Accept Rate (FAR) thresholds. Specifically, the lower the FAR, the lower the risk of the model misidentifying samples from other classes as belonging to the same class, and the higher the safety factor. The higher the TAR, the stronger the ability to correctly identify samples from the same class. The performance of ExpFace is similar to that in the six benchmark datasets. Specifically, on the CASIA training set, ExpFace still achieves performance similar to the other three margin-based softmax losses. On large-scale datasets, ExpFace achieves outstanding TAR at multiple different FAR thresholds in both IJB-B and IJB-C. Particularly, on the MS1MV3 training set, ExpFace achieves the best performance on the most representative TAR(@FAR=1e-4) in both IJB-B and IJB-C. On the WebFace4M dataset, ExpFace achieves the best performance on TAR(@FAR=1e-4) in IJB-B and the second-best performance after ArcFace on TAR(@FAR=1e-4) in IJB-C. Observing the overall performance, we can find that ExpFace can achieve more outstanding performance at lower FARs, indicating that ExpFace has a higher safety factor.

In the comparative experiments on MegaFace, we use the refined version of MegaFace \cite{deng2019arcface} for a fair evaluation, setting the number of distractors to 1 M. The final evaluation results, including rank-1 face recognition accuracy and face verification TAR at 1e-6 FAR, are shown in Tab. \ref{tb:megaface}. On the CASIA training set, ExpFace outperforms CosFace in verification accuracy. When trained on larger datasets, ExpFace shows even better performance. Specifically, on the MS1MV3 training set, ExpFace ranks second only to SphereFace in verification accuracy. On the WebFace4M training set, ExpFace ranks second only to ArcFace.

Overall, in comparison experiments with three classic margin-based softmax losses on multiple different datasets, we demonstrate that ExpFace achieves state-of-the-art accuracy.

\subsection{Exploration on Dynamic Margin Strategy}
\label{ssc:edms}
To explore the dynamic margin potential of these four margin-based softmax losses, in this section, we supervise the training of models using different margin-based softmax losses under the proposed dynamic margin strategy on the CASIA training set, and evaluate the performance of these models on various datasets. The final validation accuracies are presented in the rows corresponding to methods with the suffix "+D" in Tab. \ref{tb:comparative-soa}, Tab. \ref{tb:ijb-soa} and Tab. \ref{tb:megaface}.

The results in Tab. \ref{tb:comparative-soa} demonstrate that ExpFace enhances accuracy on the six benchmark datasets under the dynamic margin strategy, while the three classic margin-based softmax losses show accuracy fluctuations across different datasets. Tab. \ref{tb:ijb-soa} reveals that all margin-based softmax losses improve validation accuracy on IJB-B and IJB-C under this strategy. In particular, ExpFace shows the most significant accuracy improvement, with a $3.6\%$ increase on IJB-B and a $3.1\%$ increase on IJB-C. Furthermore, as observed in Tab. \ref{tb:megaface}, all margin loss except SphereFace exhibit improved performance on MegaFace. Among these, ExpFace achieves the second-highest identification accuracy, trailing only CosFace, while surpassing all other methods in verification accuracy. These results demonstrate the superior potential of ExpFace in dynamic margin strategies.

\section{Conclusion}
\label{sc:conclusion}
In this paper, we propose ExpFace, a novel margin loss for deep face recognition that uses an exponential angular margin in the angular softmax to penalize the angle between the sample feature and its corresponding class center, enhancing the discriminative power of the model. The analysis from the gradient perspective shows ExpFace effectively addresses the limitations of the three classic margin-based softmax losses and has great dynamic margin potential. Extensive experiments demonstrate ExpFace achieves state-of-the-art performance, and further exploration via the proposed dynamic margin strategy confirms its dynamic margin potential.
\begin{table}[!t] \normalsize
\centering
\renewcommand\arraystretch{1.25}
	\caption{Rank-1 face identification accuracy and face verification TAR at $10^{-6}$ FAR of different margin-based softmax losses on MegaFace}
	\label{tb:megaface}
	\begin{tabular}{l|l|c|c}
		\hline
		 \textbf{Method}&Dataset   & Id (\%)&Ver (\%)\\
   \hline
 SphereFace \cite{liu2017sphereface}&CASIA  & 81.037
&\textbf{83.036}
\\
 CosFace \cite{wang2018cosface}&CASIA  & 77.747
&76.822
\\
 ArcFace \cite{deng2022arcface}&CASIA  & 78.597
&81.586
\\
 \hline
 ExpFace (our)&CASIA  & \textbf{83.379}&82.967
\\
 \hline
 SphereFace
& MS1MV3  & 98.209&\textbf{98.595}
\\
 CosFace
& MS1MV3  & 98.337
&98.464
\\
 ArcFace
& MS1MV3  & \textbf{98.468}
&98.45
\\
\hline
 ExpFace& MS1MV3  & 98.327
&98.538
\\
 \hline
 SphereFace
& WebFace4M  & 96.431
&96.991
\\
 
CosFace& WebFace4M  & 97.595
&97.481
\\
 ArcFace& WebFace4M  & \textbf{97.597}
&\textbf{98.001}
\\
 \hline
ExpFace& WebFace4M  & 97.402
&97.896
\\
\hline

\end{tabular}
\end{table}

{
    \small
    \bibliographystyle{ieeenat_fullname}
    \bibliography{main}
}

\end{document}